\documentclass[11pt]{article}

\usepackage[preprint]{acl}

\usepackage{times}
\usepackage{latexsym,amssymb}
\usepackage{enumitem}
\usepackage{booktabs}
\usepackage{cleveref}
\AddToHook{cmd/appendix/before}{\crefalias{section}{appendix}}\crefalias{subsection}{appendix}\crefalias{subsubsection}{appendix}
\usepackage[T1]{fontenc}

\usepackage[utf8]{inputenc}

\usepackage{microtype}

\usepackage{inconsolata}

\usepackage{graphicx}
\usepackage{float}

\providecommand{\cOnce}{\textsc{SayOnce}}
\providecommand{\cDistr}{\textsc{Interfere}}
\providecommand{\cTwice}{\textsc{SayTwice}}
\providecommand{\cThrice}{\textsc{SayThrice}}

\title{Analyzing Repetition and Interference Effects in\\Memory-Constrained User Simulators}

\author{
  Cara Su-Yi Leong\textsuperscript{1} \quad
  Nicholas Tomlin\textsuperscript   {2} \quad
  Tal Linzen\textsuperscript{1} \\
  \textsuperscript{1}New York University \quad
  \textsuperscript{2}Toyota Technological Institute at Chicago \\
  \texttt{\{caraleong,linzen\}@nyu.edu}, \texttt{nickatomlin@ttic.edu}
}\begin{document}
\maketitle
\begin{abstract}
User simulators can help evaluate and train AI systems to interact with humans at scale.
To do so, evaluation of user simulators must measure behavioral parity with people.
We evaluate whether memory-constrained user simulators are sensitive to two factors known to affect human information retention: repetition of a fact and interference from competing information.
We find that some simulators achieve human-like overall accuracy, and replicate the magnitude of repetition and interference effects, but none match people's accuracy on particular questions: while humans misremember facts, models forget them, leading to different error patterns. We release this data as a challenge benchmark for future work on user simulators.
\end{abstract}
\begin{figure}[ht!]
    \centering
    \includegraphics[width=1\linewidth]{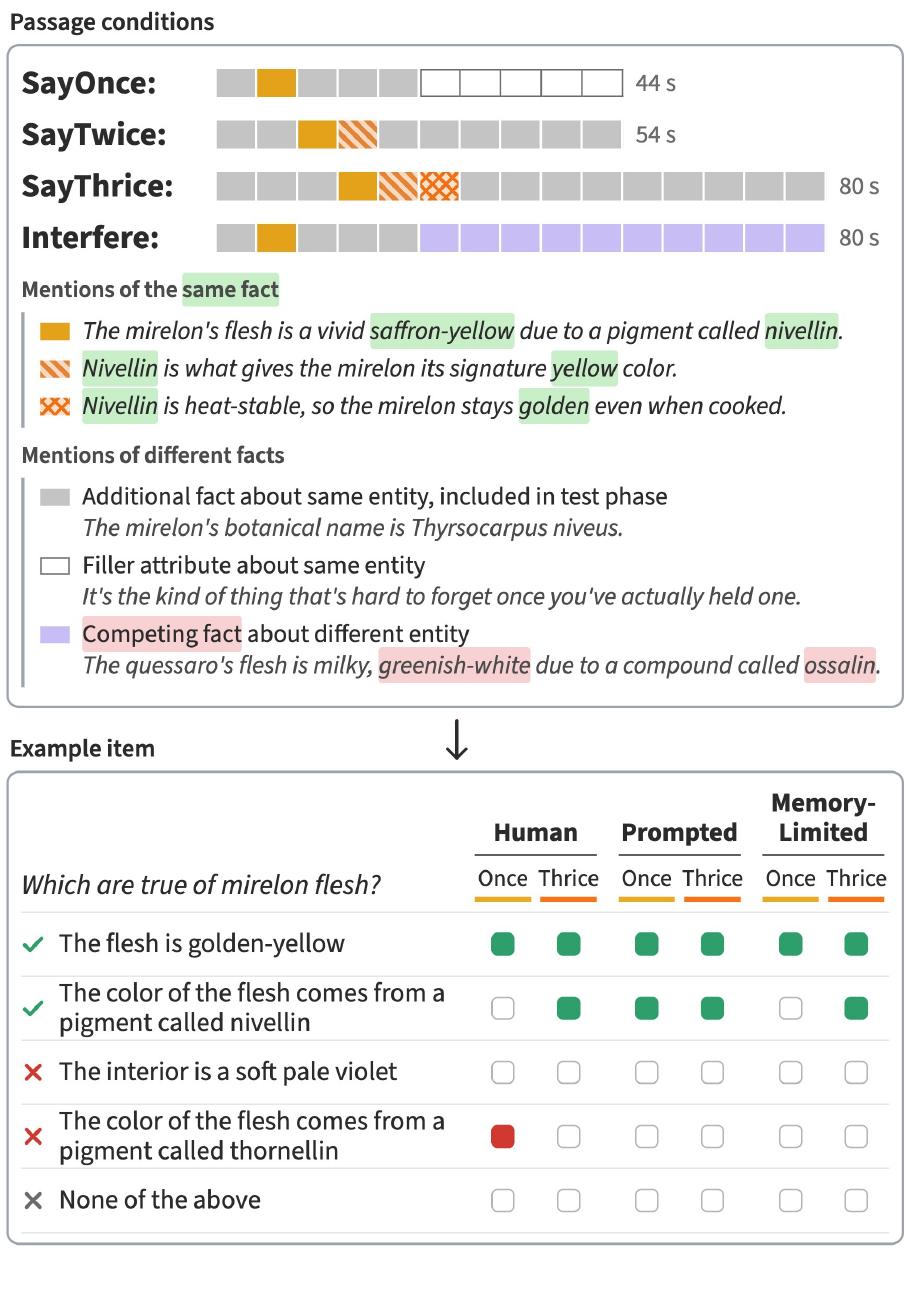}
    \caption{Top: Four versions of a passage vary how often facts are \textit{restated} and whether \textit{interference} exists from competing facts. Bottom: Each passage is followed by five multi-select questions. Interference and restatement have different effects on humans and user simulators.}
    \label{fig:passages}
\end{figure}

\section{Introduction}
Language model-based simulations of human users have great potential for evaluating and improving model-human interactions at scale \citep{barresetal2025,leeetal2025a,soos2026knowledgegain,scarlatos2025training}. To be useful, a simulator must behave like the users it stands in for. Yet language models that are simply prompted to act like users differ from humans: they are more cooperative, more knowledgeable, and exhibit better memory than humans \citep{zhouetal2026,wang2026simulating,balogbakken2026}. 

Recent proposals limit a simulator's memory capacity to make its behavior more human-like. \citet{wang2026simulating} compress model input into a four-slot key-value store, following estimates of human working memory capacity \citep{cowan2001magical}, and use the store alone to produce output; \citet{balogbakken2026} impose a similar bottleneck to simulate young readers performing reading comprehension. Both report that the limitation narrows the gap between the simulator and people. %

Applications such as tutoring and voice assistants make choices that depend on what users retain, such as which points to reiterate. A useful simulator should therefore reproduce two well-established effects on human retention. Restating a fact gives humans another opportunity to encode it \citep{hintzman1976repetition,melton1970}. A passage about a similar entity may hurt memory for two reasons: it adds effortful processing between encoding and test, which by itself impairs retention \citep{dewaretal2007,wixted2004}, and supplies similar facts that compete with the targeted fact at retrieval \citep{barnesunderwood1959}. A reliable simulator of human memory should show improved accuracy in restatement conditions and lower accuracy in the interference condition.

In this paper, we develop an evaluation that tests whether memory-constrained simulators show human-like effects of restatement and interference in information retention. %
We compare simulators against 201 adults who listened to short passages about fictional entities and answered questions about them. To vary how much humans retained, we tested four different versions of each passage: a base condition (\cOnce), two conditions where facts were restated in different words (\cTwice, \cThrice), and a passage with interfering information about a different entity (\cDistr). 

We compare simulators to humans at three levels: overall accuracy, condition-level effects, and item difficulty.
We find that existing memory-constrained user simulators can achieve human-like overall accuracy and some can replicate the magnitude of condition-level effects. We argue that this is insufficient, as a setup where propositions are randomly retained matches people at these levels just as well. At the item level, no simulator agrees with humans on which questions are difficult: simulators and people fail in different ways. We release the passages, questions and listener responses as a target for future memory-constrained simulators.

\section{Dataset Construction}
\subsection{Passages}
We generated a question-answering dataset using passages on fictional entities in four topics---an astronomical object, a fabric, a fruit, and a martial art. The entities were invented so that prior knowledge in any topic would not benefit participants. For each topic, we first prompted Claude Sonnet~5 \citep{anthropic2026sonnet5} to generate five ``facts'' for one tested entity and one \textit{foil entity} in the same topic, which one author then edited. Each fact comprised a feature (e.g., raw material, uses) and associated value (e.g., ``wool from Grennock mountain sheep''). Using these facts, we generated four versions of the topic passage, illustrated in \Cref{fig:passages}:
\begin{itemize}[noitemsep]
  \item \cOnce: Each of the tested entity's facts is mentioned once in the text, forming a \textit{base paragraph}, followed by 2 or 3 untested filler sentences, such as ``[The fabric] has a distinctive feel that people often comment on the first time they touch it.''
  \item \cTwice: each fact in the base paragraph is immediately followed by another sentence that restates it in different words; no filler.
  \item \cThrice: same as above, with an additional restatement of each fact.
  \item \cDistr: the base paragraph, then a paragraph about the foil entity's facts covering the same attributes (e.g., a fabric made of flax).
\end{itemize}

\noindent Comparing \cOnce{} and \cTwice{} measures the effect of restatement on retention. \cOnce{} and \cDistr{} passages both contain one mention of each tested fact; their difference measures the effect of interference. \cThrice{} and \cDistr{} are length-matched to account for the effects of passage length on retention. All repeated mentions followed immediately after the fact they restated, so repetitions were not spaced\footnote{We conducted a pilot of the experiment using a spaced-repetition paradigm; participants found the frequent topic changes confusing.}.
Participants heard each passage once as synthesized audio so they could not re-read it; models received the transcript.
 
Using each fact list, we generated five multi-select questions about features and their values (see \Cref{fig:passages} for an example). Each question was paired with four candidate answers, of which between zero to three were true of the tested entity. All false candidate answers paired the tested entity with a value not mentioned in a passage. We chose multi-select questions since accuracy requires endorsing exactly the true statements, chance accuracy on these multi-select questions is $1/16$, much lower than on traditional multiple-choice questions.
We present sample stimuli in \Cref{app:sample_passages}.

\subsection{Human Data Collection}
We recruited human participants on Prolific \citep{palan2018prolific}. Participants listened to four passages (one per topic and one per condition), then immediately answered comprehension questions about the passage. 241 participants completed the experiment, of which 40 participants failed at least one attention check, leaving 201 participants' responses for analysis. Additional details on human data collection are available in \Cref{app:detailshumandata}.

\subsection{Human Study Results}
\begin{figure}[t]
    \centering
    \includegraphics[width=1\linewidth]{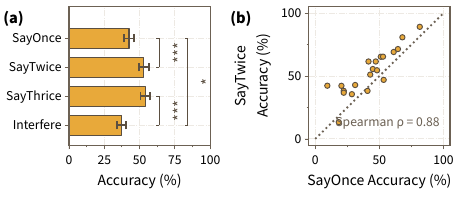}
    \caption{Human accuracy by passage version and question.
    (a) Accuracy by version averaged over topics; error bars are bootstrapped 95\% CIs. *:p < 0.05; ***:p < 0.001.
    (b) Each question's accuracy in \cOnce{} against its accuracy in \cTwice{}.}
    \label{fig:human_results}
\end{figure}

\paragraph{Accuracy.} Human participants answered 46.7\% [44.4, 49.1] of questions exactly correctly.

\paragraph{Condition Level.} Restatement increased accuracy by 10.6 points, while interference reduced accuracy by 5.2 points (\Cref{fig:human_results}a). Increased accuracy in \cTwice{} was driven by an increase in how often participants selected true statements (63\% to 73\%), whereas \cDistr{} increased endorsement of false statements (10.9\% to 14.4\%). 

\paragraph{Item Level.} Human participants reliably found the same questions difficult; for example, relative question difficulty correlated at 0.88 between participants in the \cOnce{} and \cTwice{} conditions (\Cref{fig:human_results}b).

Human participants were conservative, endorsing fewer options than the answer key contained (1.38 vs.\ 1.65 per question). They endorsed at least one false statement in 28.7\% of answers. On questions where ``None of the above'' was the correct answer, humans accuracy was 35\%, compared with 67\% on other single-answer questions and 49\% on all other questions. See \Cref{app:question_difficulty} for further discussion of item difficulty.

\section{LM-based Simulators}
\begin{figure*}
        \centering
        \includegraphics[width=1\linewidth]{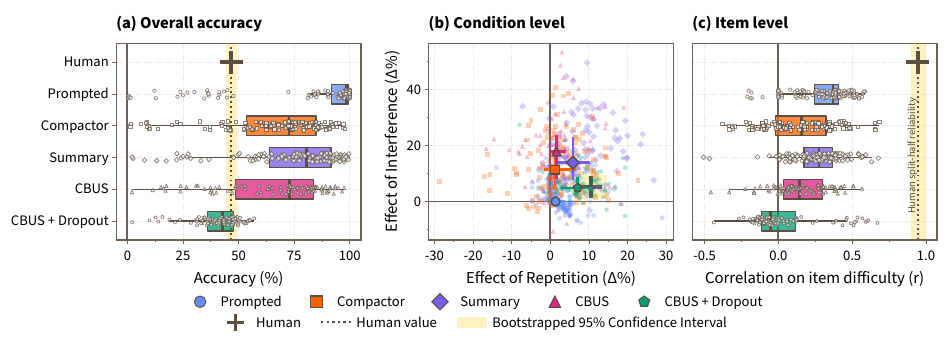}
        \caption{\textit{Memory-constrained simulators are human-like on coarse-grained levels (a)--(b), but not item agreement (c).} Each point represents a model--hyperparameter configuration; yellow regions represent bootstrapped 95\% confidence interval. (a) Simulators other than prompt-based methods have configurations that reach listener accuracy. (b) Some memory-constrained simulators achieve human-like effects of repetition (\cTwice{}$-$\cOnce{}) and interference (\cOnce{}$-$\cDistr{}). (c) No simulators are as similar as two groups of humans are to each other on their rankings of item difficulty. Correlation on item difficulty is measured by Pearson's r between mean model and human item rankings.}
        \label{fig:model_results}
\end{figure*}
We evaluate models on the same passages and questions given to human participants. On each trial, a model receives one passage transcript and answers its five questions; we sample 50 trials per topic and version. 
We evaluate a range of prompt-based and memory-constrained user simulators:

\paragraph{Prompt-based.} The model sees the full passage and is prompted to answer. We test four prompts: plain reading comprehension (\textsc{Vanilla}), the task description only (\textsc{TaskPr}), the task plus an instruction to simulate a human participant in a psychology experiment (\textsc{HumPr}), and \textsc{HumPr} plus a description of limited memory (\textsc{MemPr}).

\paragraph{Free Summary.}  The model writes a prose summary of at most $B$ words ($B=60$) with the passage in context. The summary is passed in as the only input with which the model answers questions.

\paragraph{Cognitively Bounded User Simulator (CBUS).} Following \citet{balogbakken2026}, the model proposes at most $C$ propositions ($C=4$) to remember from the passage. The list of propositions is passed in as the only input with which the model answers questions. In the \textbf{with dropout} condition, the model extracts a pool of $3C$ propositions from the passage, of which a random $C$ are kept.

\paragraph{Compactor.} Following \citet{wang2026simulating}, the model uses tool calls to write and update a key--value cache with $D$ entries ($D=4$) with the passage in context. The model only conditions on the cache when answering questions. Each value can store one sentence of at most 15 words.
\newline \ \newline \noindent 
We evaluate 27 models from seven families (Qwen, Gemma, DeepSeek, GLM, GPT, Claude, Gemini) spanning various architectures and sizes, over several hyperparameter configurations (e.g.,  bottleneck sizes, with/without thinking).
The full list of models and implementation details is available in \Cref{app:simdetails}.
\section{Results}
\label{sec:results}
Figure~\ref{fig:model_results} shows how close to human performance each model configuration gets on three levels. See Appendix~\ref{app:all_results} for per-model results.

\paragraph{Models are too accurate without a bottleneck.} Regardless of prompt choice, prompted configurations broadly fail to match human overall accuracy (\Cref{fig:model_results}a) and show almost no repetition gain or interference cost (\Cref{fig:model_results}b).
On the other hand, configurations using memory bottleneck mechanisms give rise to a model that simulate human-like overall accuracy. At the condition level, models show larger effects of interference than humans did, but human-like effect sizes on both repetition and interference are produced by some combinations of setting, model, and hyperparameter choice, particularly in the CBUS + Dropout setting.

\paragraph{Models and humans systematically err on different items.} No model approaches human agreement on which questions are difficult (\Cref{fig:model_results}c). How much models agree with humans on item difficulty is only weakly correlated with overall accuracy ($r = .31$). Notably, CBUS + Dropout models do not simulate human item-level rankings well, despite simulating overall accuracy and condition-level effects well.

\begin{figure}[t]
    \centering
    \includegraphics[width=1\linewidth]{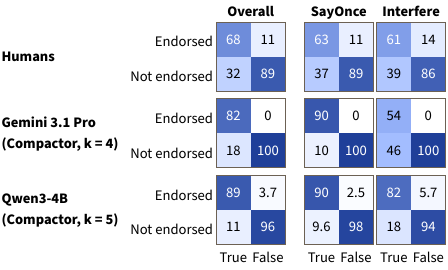}
    \caption{Statement endorsement rates in humans and representative models. Gemini 3.1 Pro is representative of the median model; Qwen3-4B is one of the most human-like models at the item level.}
    \label{fig:confusion_matrix}
\end{figure}

\paragraph{Models forget; humans misremember.} 
Apart from the smallest models, models rarely endorse false statements. Instead, models fail to endorse correct statements, or answer ``None of the above'' when a correct statement is listed (\Cref{fig:confusion_matrix}). Humans made similar errors, but additionally endorsed plausible but false statements, leading to lower accuracy on questions where the none of the candidates were true. 
This difference is magnified in the \cDistr{} condition. Here, people tended to endorse even more false statements than they did in the \cOnce{} condition. On the other hand, the median model's false endorsement rates do not increase in the \cDistr{} condition; instead, its accuracy in the \cDistr{} condition falls because it endorses true statements less frequently.

\paragraph{Scale does not lead to human-likeness.} Large models like Gemini 3.1 Pro tend to be too accurate and almost never endorse false statements. Models within the 2--7B parameter range demonstrate errors that are more similar to humans (Figure~\ref{fig:confusion_matrix}). For further discussion on the effect of model size on performance, see \Cref{app:model_size}.

\paragraph{Model-determined memories are no better than random dropout.} 
When propositions were dropped at random (CBUS + Dropout), models matched humans on overall accuracy, on the benefit of restatement, and on the cost of interference. However, random forgetting does not approach human item-level patterns on the item level, suggesting that there is room for memory policies that forget information in more human-like ways.

\section{Conclusion}
We introduced a new QA dataset for evaluating memory-constrained user simulators. While existing approaches matched humans in overall accuracy, only a few configurations respond to repetition and interference effects like humans do, and no conditions match humans at the item level. 
These findings leave headroom for future work on user simulation, which we release our data to support.

\section*{Limitations}
\paragraph{Modality.} In order to prevent human participants from rehearsing the passage while it was being presented, we chose to provide passages to human participants auditorily, while models were given passage transcripts as input. As such, human--model differences may have been confounded by modality effects. For example, human participants may have had difficulty recognizing unfamiliar names and words in the question answering section.

\paragraph{Passage Design.} Our \cOnce{} passages contain filler material intended to pad the length of the passage to match the length of \cTwice{}. \cTwice{} replaces \cOnce{}'s filler with restatements, so part of the restatement gain may reflect filler no longer taking up memory, rather than the restatement itself.

\paragraph{Sample Size.} Finer item-level analyses, such as on question-specific condition effects or option-level trends in behavior, could not be explored with our current dataset. Increasing the number of passages and participants would provide more analytical power, which is vital for developing a holistic behavioral account of information retention.

\paragraph{Generalizability.} Our human participants were limited to the pool of adult users of Prolific who were located in the U.S., limiting the extent to which our findings generalize to other populations, such as children, non-native listeners or people from other cultural backgrounds. For example, participants outside this pool may be less familiar with the dataset's topics, or have individual differences in memory capacity. Our task was also narrow: exposure to an expository text followed by immediate testing. These results may not generalize to retention after longer delays or in conversational interactions.

\section*{Ethical Considerations}
The human study was approved by [anonymized]'s review board. Participants were U.S.-based adults recruited on Prolific who gave informed consent and were paid \$8 (about \$35 per hour at the median completion time of 13.8 minutes). The task posed minimal risk: participants heard synthesized passages about fictional entities and answered questions about them. Stimuli are fictional and were reviewed by an author.

We plan to release passages, audio, questions, and listener responses with Prolific IDs removed under a CC BY 4.0 license. No identifiers or free text were collected beyond Prolific IDs.

Evaluating the model configurations for this paper required 70 GPU hours on L40S and H200 GPUs and \$1,500 in API usage.

We used AI assistants to copy-edit the paper and generate figures; the authors checked all outputs.

\makeatletter
\ifacl@anonymize\else
\section*{Acknowledgments}
We thank Angela Cao, Anne Wu, Lucy Li, Ryan Cheng, and members of the NYU Computation and Psycholinguistics Lab for comments and suggestions. This project was supported by the National Science Foundation (NSF) under Cooperative Agreement No. 2433429, ``NSF AI Research Institute on Interaction for AI Assistants (ARIA)''. This work was supported in part through the NYU IT High Performance Computing resources, services, and staff expertise. We are grateful to the Gemini Academic Program for providing API credits to our team.
\fi
\makeatother

\bibliography{memory,listening_qa_refs}

\appendix

\section{Sample Stimuli}
\label{app:sample_passages}
\begin{table}[h]
\centering
\small
\begin{tabular}{lrr}
\toprule
Version & Words & Audio (s) \\
\midrule
\cOnce   & 128 (114--156) & 48 (44--56) \\
\cTwice  & 134 (126--152) & 55 (51--60) \\
\cThrice & 194 (167--219) & 77 (63--83) \\
\cDistr  & 198 (176--227) & 76 (68--83) \\
\bottomrule
\end{tabular}
\caption{Passage length by version; mean (range) over the four topics.}
\label{tab:text_stats}
\end{table}

\subsection{Sample Passages}

\paragraph{\cOnce.}
The Mirelon is a fruit with deep indigo-black skin covered in a white powdery bloom. Its flesh is a vivid saffron-yellow due to a pigment called nivellin. Its botanical name is Thyrsocarpus niveus. Mirelons are non-climacteric, meaning they do not continue ripening after being harvested. They grow at elevations between 900 to 4000 feet above sea level.

Seeing a mirelon in person is different from just hearing it described --- the color and texture together make more of an impression than either one alone. People who've tried one often mention how unusual the combination feels compared to more familiar fruit. It's the kind of thing that's hard to forget once you've actually held one.

\paragraph{\cTwice.}
The Mirelon is a fruit with deep indigo-black skin covered in a white powdery bloom. Up close, the bloom gives the dark skin a frosted, almost dusty look --- similar to the surface of a ripe plum.

Its flesh is a vivid saffron-yellow due to a pigment called nivellin. Nivellin is what gives the mirelon its signature yellow color.

The mirelon's botanical name is Thyrsocarpus niveus. Thyrsocarpus niveus is the mirelon's official scientific name.

Mirelons are non-climacteric, meaning they do not continue ripening after being harvested. Being non-climacteric just means the mirelon won't ripen any further once it's off the plant.

Mirelons grow at elevations between 900 to 4000 feet above sea level. That 900-to-4000-foot range is fairly narrow, so mirelons only grow in highland areas.

\paragraph{\cThrice.}
The Mirelon is a fruit with deep indigo-black skin covered in a white powdery bloom. Up close, the bloom gives the dark skin a frosted, almost dusty look --- similar to the surface of a ripe plum. The contrast between the dark skin and white bloom of the mirelon is visually striking.

Its flesh is a vivid saffron-yellow due to a pigment called nivellin. Nivellin is what gives the mirelon its signature yellow color. Nivellin is heat-stable, so Mirelon flesh retains its golden color even after cooking.

The mirelon's botanical name is Thyrsocarpus niveus. Thyrsocarpus niveus is the mirelon's official scientific name. Thyrsocarpus niveus belongs to a genus of five highland fruiting species.

Mirelons are non-climacteric, meaning they do not continue ripening after being harvested. Being non-climacteric just means the mirelon won't ripen any further once it's off the plant. Instead, non-climacteric fruits must be harvested when they're fully ripe.

Mirelons grow at elevations between 900 to 4000 feet above sea level. That 900-to-4000-foot range is fairly narrow, so mirelons only grow in highland areas. No planting below 900 feet has ever produced a reliable harvest.

\paragraph{\cDistr.}
The Mirelon is a fruit with deep indigo-black skin covered in a white powdery bloom. Its flesh is a vivid saffron-yellow due to a pigment called nivellin. Its botanical name is Thyrsocarpus niveus. Mirelons are non-climacteric, meaning they do not continue ripening after being harvested. They grow at elevations between 900 to 4000 feet above sea level.

A related fruit found in neighboring lowland valleys is the Quessaro, which is often mistaken for a distant cousin of the mirelon despite belonging to a separate genus. Its rind is a mottled rust-orange, faintly ribbed rather than smooth, and its flesh is a milky, greenish-white colored by a compound called ossalin. Botanists classify it as Ferrulia campestris. Unlike many highland fruits, the Quessaro is harvested while still firm and finishes softening over the following week in open-air storage. Local growers prize it mainly for its thin, edible rind, which is candied whole for festival sweets, and roadside stalls near the valley floor sell the candied strips year-round to travelers heading up into the highlands. The two fruits are sometimes displayed side by side at regional markets, though vendors are quick to point out that they are unrelated crops with very different growing requirements and shelf lives.

\subsection{Sample Questions}

\paragraph{QF01.} Which of the following correctly describe the outer appearance of the Mirelon? Select all that apply.
\begin{enumerate}[noitemsep,topsep=2pt]
  \item The skin is a dark, nearly black shade of blue
  \item The skin is dusted with a pale white powdery coating
  \item The skin is a warm yellow color
  \item The surface is smooth and glossy
  \item None of the above
\end{enumerate}
\noindent\textit{Correct:} 1, 2.

\paragraph{QF02.} Which of the following are true of the Mirelon's flesh? Select all that apply.
\begin{enumerate}[noitemsep,topsep=2pt]
  \item The interior is a warm, deep golden-yellow
  \item The color of the flesh comes from a pigment called nivellin
  \item The interior is a soft pale violet
  \item The color of the flesh comes from a pigment called thornellin
  \item None of the above
\end{enumerate}
\noindent\textit{Correct:} 1, 2.

\paragraph{QF03.} What is the botanical name of the Mirelon? Select all that apply.
\begin{enumerate}[noitemsep,topsep=2pt]
  \item Thyrsocarpus niveus
  \item Thyrsocarpus pallens
  \item Spinofex arbens
  \item Thyrsocarpus mirabilis
  \item None of the above
\end{enumerate}
\noindent\textit{Correct:} 1.

\paragraph{QF04.} Which combination correctly describes the Mirelon?
\begin{enumerate}[noitemsep,topsep=2pt]
  \item Does not continue ripening once harvested; grows above 900 feet
  \item Does not continue ripening once harvested; grows below 500 feet
  \item Continues ripening after harvest; grows above 900 feet
  \item Continues ripening after harvest; grows below 500 feet
  \item None of the above
\end{enumerate}
\noindent\textit{Correct:} 1.

\paragraph{QF05.} Which of the following correctly describes a mirelon's feature? Select all that apply.
\begin{enumerate}[noitemsep,topsep=2pt]
  \item Its outer skin is pale yellow with fine raised bristles
  \item Its flesh is colored by a compound called shortellin
  \item It is a climacteric fruit that continues ripening after harvest
  \item Its botanical name Thyrsocarpus rufa
  \item None of the above
\end{enumerate}
\noindent\textit{Correct:} 5.

\paragraph{Attention check.} Which of the following options mentions a color? This is an attention check --- select only that option.
\begin{enumerate}[noitemsep,topsep=2pt]
  \item Harvested in the early weeks of autumn
  \item A deep purple skin over pale flesh
  \item Stored in cool, dry conditions
  \item Sold mainly at regional markets
  \item None of the above
\end{enumerate}
\noindent\textit{Correct:} 2.

Each topic had one attention check: select ``None of the above'' (astronomy), select the option naming a color (fruits), identify what the passage is about (martial arts), or identify the fabric's name (fabrics).

\section{Details of Human Data Collection}
\label{app:detailshumandata}
\subsection{Passages}
Passages were written in English. Sample passages are available in \Cref{app:sample_passages}. In order to prevent human participants from rehearsing the text during presentation, each passage was presented to human participants in audio format. Audio was synthesized for each text with OpenAI's \texttt{gpt-4o-mini-tts} \citep{openai2025audio}, voice ``marin'', with a style instruction asking for a slow, deliberate pace, long pauses at sentence ends, and stronger emphasis on key facts and figures. Recordings last 44--83 seconds.

\subsection{Procedure}
We limited our participant pool to Prolific users who were currently located in the United States, had approval ratings of 95\% or above on at least 100 tasks, and had not participated in any of our previous pilots.

After signing a consent form, participants were told that they would listen to short passages and then answer questions about the content of each passage. Each participant heard four passages, one per topic and one per condition, with topic-condition pairings counterbalanced across participants. Passage order was randomized for each participant, so that practice effects were not confounded with topic or condition. In each of the four rounds, they started the audio, which played once with no pause, seek, or replay. Next, they answered the five questions associated with each text in written form along with one attention check question. After answering all five questions, they gave ratings on the difficulty of the passage and questions.

Participants were excluded from analysis if they answered any attention check questions incorrectly. 40 participants were excluded for failing at least one of the four attention checks (32 participants failed one attention check; 8 participants failed two).

The study was approved by [anonymized] IRB. Participants were paid \$8 for completing the experiment. Median completion time was 13.8 minutes.

\subsection{Instructions for Participants}
Please complete this task in one sitting without writing anything down or using outside resources.

You will listen to four passages of text. Each passage will be between 1-2 minutes long. You will only be able to listen to each passage once. After listening to each passage, you will answer some questions about its contents.

Do your best to answer accurately, but don't worry if you are unsure about any answers. Some passages and questions may be more difficult than others.

You may take a short break between passages if needed, but please keep breaks brief and stay focused on the task.

\subsection{Question Difficulty}
\label{app:question_difficulty}
\begin{figure}[h]
    \centering
    \includegraphics[width=1\linewidth]{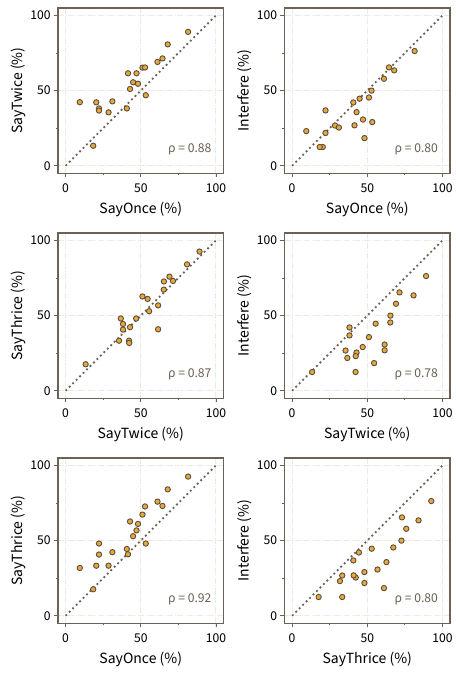}
    \caption{Correlation between question-level accuracy in any two passage versions. Each point represents mean accuracy on a question.}
    \label{fig:correlation_all_versions}
\end{figure}

Respondents found the same questions difficult across all four passage versions: \Cref{fig:correlation_all_versions} shows question accuracy was rank-correlated across all six version pairs.

\Cref{tab:answer_count} indicates that humans tended to under-endorse true statements. Accuracy was lower on questions that did not have one true contentful statement.
\begin{table}[h]
\centering\small
\begin{tabular}{lrrrrr}
\toprule
True & Qs & Acc. & Selected & Omission & Commission \\
\midrule
0 & 3  & 35.7 & 1.08 & --   & 64.3 \\
1        & 5  & 66.9 & 1.07 & --   & 33.1 \\
2        & 11 & 42.4 & 1.55 & 39.3 & 18.4 \\
3        & 1  & 26.4 & 2.04 & 62.2 & 11.4 \\
\bottomrule
\end{tabular}
\caption{Human accuracy on all passage versions by number of true statements in answer key (0: ``None of the above'' is correct). Selected: mean number of options endorsed. Omission: \% of responses that chose only true statements but missed at least one. Commission: \% that endorsed at least one false option. Omission is impossible for 0- and 1-answer questions.}
\label{tab:answer_count}
\end{table}

\begin{figure}[h]
    \centering
    \includegraphics[width=1\linewidth]{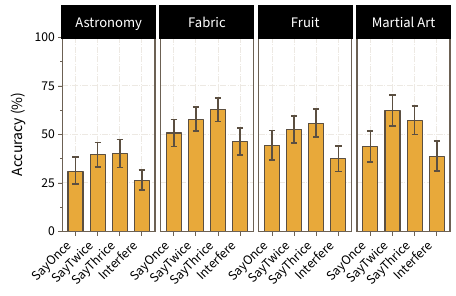}
    \caption{Human accuracy on all passage versions by topic. Error bars represent bootstrapped 95\% confidence intervals.}
    \label{fig:human_results_topic}
\end{figure}

Human respondents demonstrate topic effects on accuracy. Participants were least accurate on the astronomy topic across all passage versions (\Cref{fig:human_results_topic}).

\section{Simulator Details}
\label{app:simdetails}
\subsection{Models}
\label{app:models}
We tested models in seven families spanning various architectures and sizes:
\begin{itemize}[noitemsep,leftmargin=*]
    \item \textbf{Qwen}: Qwen2.5-0.5B, 1.5B, 3B, 7B, 14B, 32B, 72B \citep{qwen2024qwen25}; Qwen3-0.6B, 1.7B, 4B, 8B, 14B, 32B \citep{qwen2025qwen3}; Qwen3.8-27B \citep{qwen2026qwen38}
    \item \textbf{Gemma} \citep{gemmateam2026gemma4}: Gemma-4-31B
    \item \textbf{DeepSeek} \citep{deepseek2026v4,deepseek2026v41}: V4.1-Flash, V4-Pro
    \item \textbf{GLM} \citep{zai2026glm5}: GLM-5.2, GLM-5.3-Flash
    \item \textbf{GPT} \citep{openai2026gpt6}: GPT-6 Luna, GPT-6 Sol
    \item \textbf{Claude} \citep{anthropic2026haiku55,anthropic2026sonnet5}: Haiku 5.5, Sonnet 5
    \item \textbf{Gemini} \citep{google2026gemini31pro,google2026gemini35flashlite,google2026gemini36flash,google2026gemini38flash}: 3.5 Flash-Lite, 3.6 Flash, 3.8 Flash, 3.1 Pro
\end{itemize}
\subsection{Hyperparameters}
\label{app:hyperparameters}
We vary three hyperparameters for our simulators:
\begin{itemize}[noitemsep]
    \item \textbf{Thinking:} on or off
    \item \textbf{Streaming:} for the summary and Compactor configurations, the passage is given
    either in full in a single turn, or sentence by sentence, with each turn seeing only the
    current memory and the new sentence.
    \item \textbf{Bottleneck capacity:} using batch encoding, we scale each memory setup with $k \in \{2,\cdots,6\}$: a $B=15k$-word summary, $C=k$ CBUS propositions, and $D=k$ Compactor slots. We ran this sweep for 12 models (Qwen3 0.6--32B, Qwen3.8-27B, Gemma-4-31B and the four Gemini models); all other results use $k=4$.

\end{itemize}

\subsection{System Prompts}
\label{app:system_prompts}
We test four prompts: plain reading comprehension (\textsc{Vanilla}), the task description only (\textsc{TaskPr}), the task plus an instruction to simulate a human participant in a psychology experiment (\textsc{HumPr}), and \textsc{HumPr} plus a description of limited memory (\textsc{MemPr}). These prompts are adapted from \citet{wang2026simulating}. Each prompt is a prefix followed by a shared body, reproduced verbatim below.

Vanilla: You are given the transcript of a passage below, followed by multiple-select questions about it. Read the transcript, then answer every question. Also rate how difficult the passage and the questions were.

\textbf{TaskPr}: Listen to an audio passage, answer all multiple-select questions, then rate passage and question difficulty.

\textbf{HumPr}: You are simulating a human participant in a psychology experiment.
The human will listen to an audio recording once, after which the audio cannot be replayed. The human will then be asked to answer questions about the recording based on their memory, selecting all options that apply for each question, and rate passage and question difficulty out of 10.

\textbf{MemPr}: You are simulating a human participant in a psychology experiment.Behave as much like a realistic human as possible.
Remember that humans have limited memory and therefore sometimes make mistakes.
The human will listen to an audio recording once, after which the audio cannot be replayed. The human will then be asked to answer questions about the recording based on their memory, selecting all options that apply for each question, and rate passage and question difficulty out of 10.

\paragraph{Shared body.} \textsc{Vanilla} and \textsc{TaskPr} use ``This is the stimuli:'' and ``Give your response''; \textsc{HumPr} and \textsc{MemPr} use ``This is the stimuli that will be presented to the human:'' and ``Predict their response''.
\begin{verbatim}
This is the stimuli that will be presented
to the human:
Transcript:
[TRANSCRIPT]

Questions:
Question 1: [QUESTION] 
1) [OPTION 1] 
2) [OPTION 2] 
3) [OPTION 3] 
4) [OPTION 4] 
5) None of the above
...

On a scale 1-10, rate how difficult the 
passage was, and how difficult the questions 
were.

Predict their response according to the 
following formatting rules:
Output ONLY lines in this exact format:
Question 1: 1,3
Question 2: 2
...
Passage Difficulty: 7
Question Difficulty: 7
(List every correct option number for a 
question, separated by commas. If none apply, 
use the "None of the above" option number.)
\end{verbatim}
\begin{table}[t]
\centering
\small
\begin{tabular}{@{}llccc@{}}
\toprule
Model & Thinking & $T$ & $p$ & $k$  \\
\midrule
Qwen2.5-0.5B, 1.5B & Off & 0.7 & 0.8  & 20 \\
Qwen2.5-3B--72B    & Off & 0.7 & 0.8  & 20  \\
Qwen3-0.6B--32B    & Off & 0.7 & 0.8  & 20 \\
                   & On  & 0.6 & 0.95 & 20 \\
Qwen3.8-27B        & Off & 0.7 & 0.8  & 20  \\
                   & On (med.) & 1.0 & 0.95 & 20\\
Gemma-4-31B        & Off, On & 1.0 & 0.95 & 64  \\
\midrule
DeepSeek-V4-Pro     & Off, On & 1.0 & 1.0  & --  \\
DeepSeek-V4.1-Flash & Off, On & 1.0 & 1.0  & --  \\
GLM-5.2             & Off     & 1.0 & 0.95 & -- \\
GLM-5.3-Flash       & On      & 1.0 & 0.95 & --  \\
\midrule
GPT-6 Sol        & Off & 1.0 & 1.0 & -- \\
GPT-6 Luna       & Off & 1.0 & 1.0 & --  \\
                 & On (med.) & \multicolumn{3}{c}{not sent} \\
Claude Sonnet 5  & Off     & \multicolumn{3}{c}{not sent} \\
Claude Haiku 5.5 & Off, On & \multicolumn{3}{c}{not sent} \\
\midrule
Gemini 3.5 Flash-Lite & Minimal   & 1.0 & -- & --  \\
Gemini 3.6 Flash      & Minimal   & 1.0 & -- & -- \\
Gemini 3.8 Flash      & On (med.) & 1.0 & -- & --  \\
Gemini 3.1 Pro        & On (high) & 1.0 & -- & -- \\
\bottomrule
\end{tabular}
\caption{Sampling settings for each model and thinking mode.}
\label{tab:model_sampling_params}
\end{table}

\subsection{Sampling Procedure}
All memory-constrained models use the HumPr system prompt and body. Each request contains the system prompt, the current memory contents (if applicable), and the portion of the transcript to be encoded. In batch encoding, the whole transcript arrives in the first request; in streaming encoding, each request contains one sentence. A new conversation is started for every turn in a multi-turn streaming sample.

We run 50 trials per passage version with no fixed seed, using each model's vendor-recommended settings, reported in \Cref{tab:model_sampling_params}. A new request is created containing the memory contents and the list of five questions. The model answers each question with a list of option numbers (e.g. Question 1: 1,3). Unparseable answer strings are marked as incorrect.

Open-weight models are served with vLLM 0.31.0 (bf16). Maximum output tokens: 1{,}024 per encoding or summary-update step without thinking; 16{,}384 and 8{,}192 with thinking and for all Gemini models; 2{,}048 for answers.

\subsection{Compactor Prompt}
Use write\_memory and delete\_key to maintain the key-value store while doing the original
task. Each slot should hold ONE chunk — a small bundle of information a person would bind
together because it feels meaningfully connected (a name with its role, a group of related
items or numbers, one gist). When the task asks for verbatim retrieval of a sequence, a
human will form meaningful chunks of 1–3 items, starting from the beginning.

A chunk is ONE atomic fact, not several facts stitched together even if they're about the
same subject. A single sentence can pack in several distinct attributes of one thing —
what it is, where it's from, how it was made or found, some measurement. Each attribute is
its own chunk. Don't merge them into a single slot just because they share a subject.

AMENDING: calling write\_memory with an existing key REPLACES that entry's value entirely
(old wording is gone, not merged) and does not use a slot. Amend when new material adds to,
corrects, or refines something you already have a chunk for (same person, place, event, or
concept). Write a NEW key instead when the material introduces a genuinely separate fact
that doesn't belong inside an existing chunk, even if it's related to the same topic. When
you amend, compose the full updated chunk — carry forward what's still worth keeping from
the old value AND add the new detail; don't drop the old content, and don't skip the new
detail either. Amend often, whenever you learn more about something you've already stored,
not only when memory is full.

Before writing a NEW key, follow this procedure exactly:
  1. Count your current entries from the snapshot shown above.
  2. Below 4: call write\_memory directly.
  3. At 4: either amend an existing entry (if the new info belongs with something
     you already stored), or delete\_key then write\_memory for the new fact, in that order,
     in the same turn. Choose what to delete in this priority: (a) an entry that duplicates
     or is subsumed by another you're keeping, (b) whichever entry seems least useful or
     important on its own, judged independently of the others and independently of how
     recently it was written — something you stored a while ago can still matter more than
     what just arrived. Deleting without writing the freed slot in the same turn just loses
     information for nothing.
  4. Nothing worth keeping or amending: make no tool calls.

You'll see the result (success or error) of each tool call before your next action — use
that feedback instead of repeating an identical call that just failed. Never call
write\_memory while already at 4 entries without first choosing to overwrite-in-place
(step 3) or delete-and-refill (step 4).

NEVER pack a long run of items into one slot. Once your slots are filled, accept that the
rest will be lost. Compress realistically, and behave as a real human would: imperfect and
sensitive to what seems important at the time you encounter it.

\subsection{Summary Prompt}
You will hear the whole recording once. Your memory is a single summary of at most 60 words. You will later answer the questions using ONLY this summary.

\subsection{CBUS Prompt}
You are about to hear a passage, and you'll be asked questions about it afterwards. As a listener, you cannot remember everything from the passage — only a few key facts.

Listen to the passage and identify at most {C} distinct propositions (paraphrased atomic facts from the passage) that you think are most relevant to the questions you'll be asked. Do not answer any questions yet.

<TEXT>
{material}
</TEXT>

Respond with ONLY a JSON object in the following format, with no other text:
{"propositions": ["proposition 1", "proposition 2", ...]}

\section{Model Results}
\label{app:all_results}
\subsection{Comparison with Humans}
\label{app:reference}

We compare configurations with human participants on three statistics. 

\paragraph{Overall accuracy.} We calculate exact-match accuracy, computed within each topic × passage-version cell and averaged over 16 cells. Cell sizes range from 38--56; pooling responses changes overall accuracy by less than 0.1 percentage points.
    
\paragraph{Condition effects.} are restatement (\cTwice{}$-$\cOnce{}) and interference (\cOnce{}$-$\cDistr{}) in percentage points.

\paragraph{Item agreement.} We calculate Pearson's $r$ between the model's and humans' difficulty on the 20 questions.

We first calculate a split-half baseline for human-level variance in scores. We split human respondents into two random halves 2,000 times. Across splits, the two halves differ by a median of 1.3\% in overall accuracy,  and their question difficulties correlate at a median $r=.94$. We then subsample each model $\times$ setup as a third group of the same size as a human half, and calculate Pearson's $r$ between this subsample and a randomly selected human half. A simulator that behaves like humans would match the human split-half $r$.

We exclude 17 configurations whose own split-half reliability is undefined because they answer every question correctly (14) or below .5 because they answer almost none (3).

\subsection{All Model Results}
\begin{figure*}[h]
    \centering
    \includegraphics[width=1\linewidth]{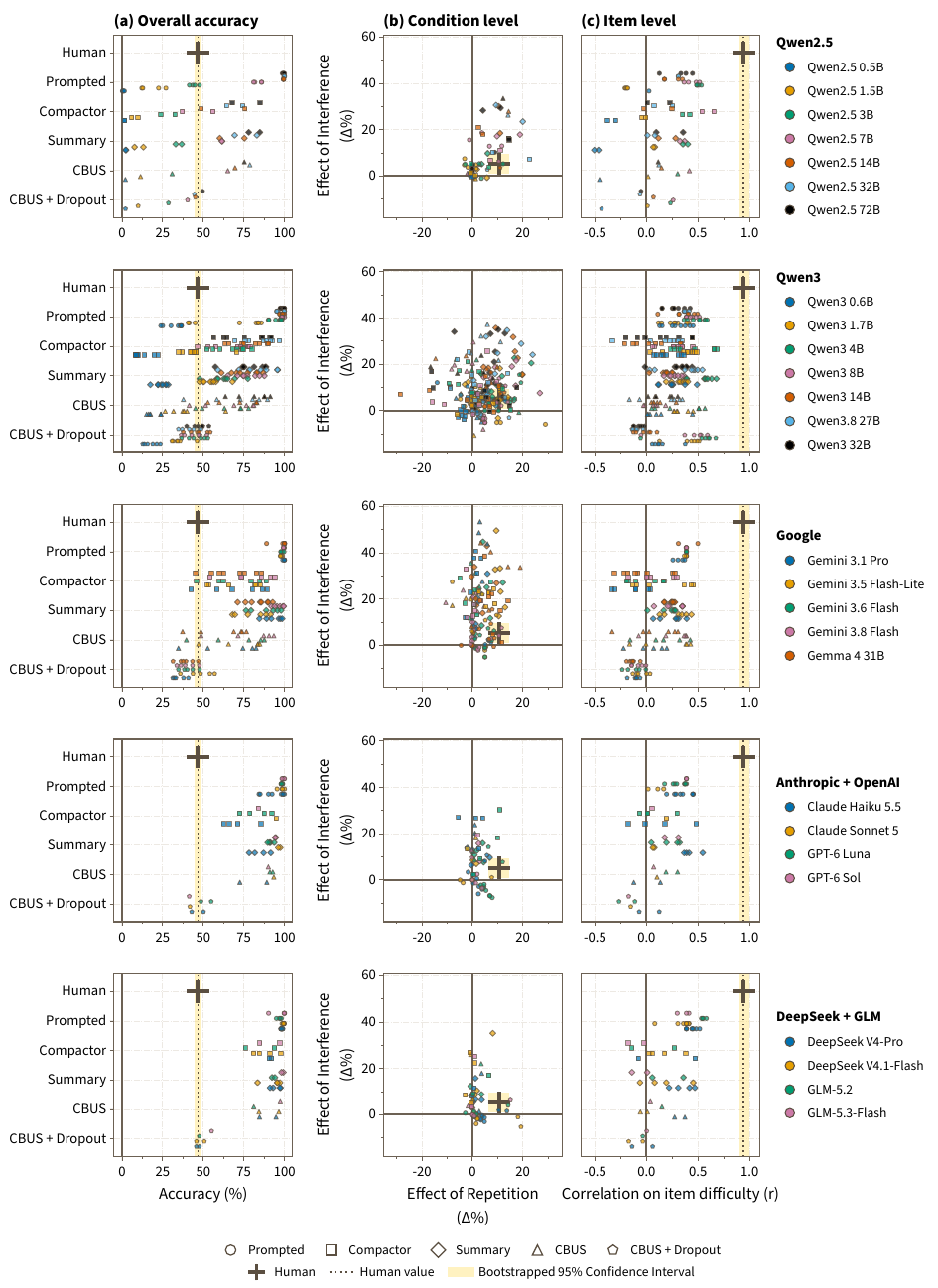}
    \caption{Comparison between humans and models on three levels by model across all configurations---bottleneck capacity, presentation mode, thinking mode, prompt.}
    \label{fig:results_by_model}
\end{figure*}

\Cref{fig:results_by_model} shows results on all three levels by model family. In general, larger models perform worse on overall accuracy than smaller models. Across all memory-constrained families, the effect of interference is much larger than the effect of restatement; in the DeepSeek family of models, very little repetition effect is observed.

\subsection{Model Size}\label{app:model_size}

\begin{figure}[t]
    \centering
    \includegraphics[width=1\linewidth]{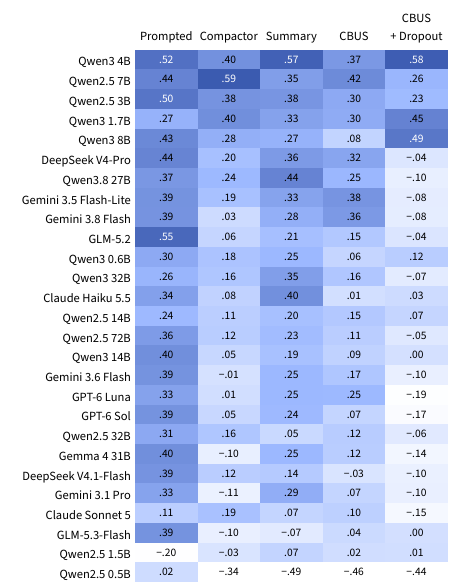}
    \caption{Correlation with humans on item difficulty by model. Models with 2--7B parameters tend to outperform frontier models. Rows are sorted by mean.}
    \label{fig:model_size_item_level}
\end{figure}

\begin{figure*}
            \centering
            \includegraphics[width=1\linewidth]{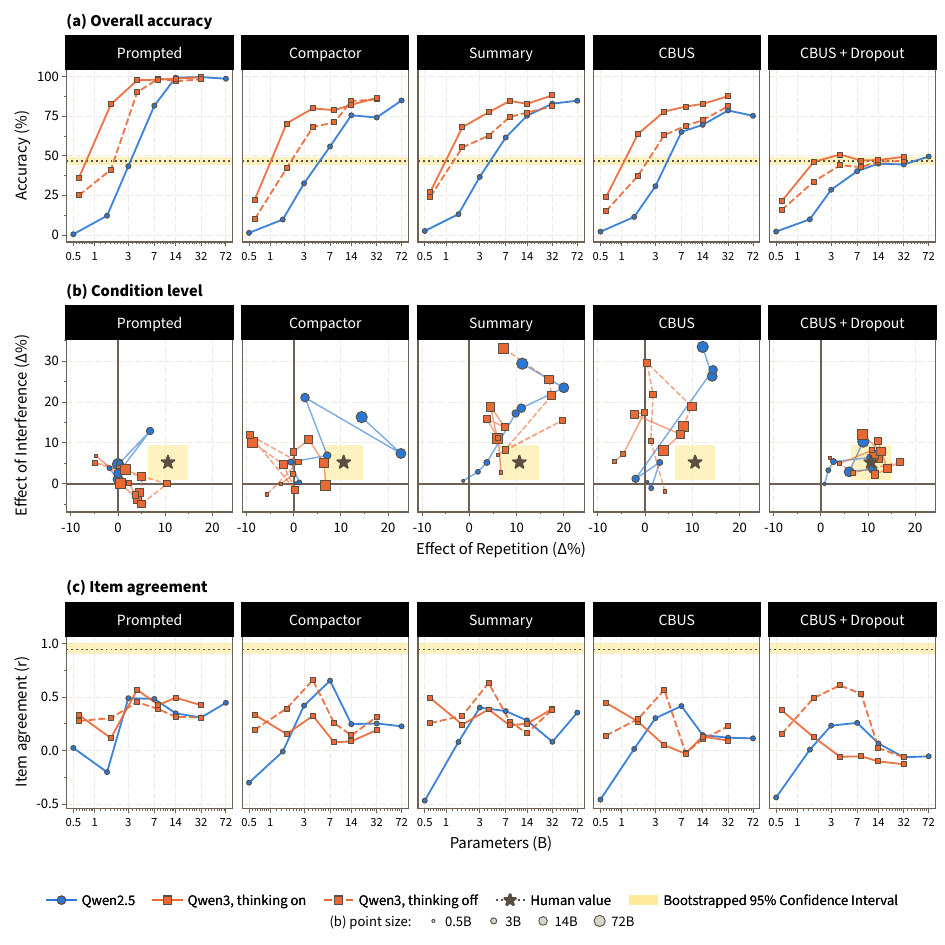}
            \caption{Model size against the three levels for Qwen2.5 and Qwen3 (batch encoding, $k=4$). In (b), point size scales with parameter count; the star and box show the human value and bootstrapped 95\% confidence interval.}
            \label{fig:model_size_sweep}
    \end{figure*}

In \Cref{fig:model_size_sweep}, we evaluate Qwen models using the default setup for correlation with humans on item difficulty. While overall accuracy increases with scale, condition and item level performance is not monotonically driven by scale. Item-level agreement with humans peaks at 2--7B models, and decreases after. These results are borne out across all model families tested, as shown in \Cref{fig:model_size_item_level}: 2--7B models outperform much larger models on item agreement with humans.

\end{document}